# Artificial Intelligence for Secured Information Systems in Smart Cities: Collaborative IoT Computing with Deep Reinforcement Learning and Blockchain

Amin Zakaie Far[1], Mohammad Zakaie Far[1], Sonia Gharibzadeh[2], Hajar Kazemi Naeini[3], Leila Amini[4], Shiva Zangeneh[5], Morteza Rahimi[6,*], Saeed Asadi[3]

1- Department of Mechanical Engineering, Politecnico di Milano, Milano, Italy
2- Department of Computer Engineering, South Tehran Branch, Islamic Azad University, Tehran, Iran
3- Department of Civil Engineering, University of Texas at Arlington, Arlington, Texas
4- Department of Information Systems and Business Analytics, Florida International University, Miami, FL, USA
5- Faculty of Engineering, University of Malayer, Malayer, Iran
6- School of Computing and Information Sciences, Florida International University, Miami, FL, USA
Corresponding author: mrahi011@fiu.edu

**Abstract**

The accelerated expansion of the Internet of Things (IoT) has raised critical challenges associated with privacy, security, and data integrity, specifically in infrastructures such as smart cities or smart manufacturing. Blockchain technology provides immutable, scalable, and decentralized solutions to address these challenges, and integrating deep reinforcement learning (DRL) into the IoT environment offers enhanced adaptability and decision-making. This paper investigates the integration of blockchain and DRL to optimize mobile transmission and secure data exchange in IoT-assisted smart cities. Through the clustering and categorization of IoT application systems, the combination of DRL and blockchain is shown to enhance the performance of IoT networks by maintaining privacy and security. Based on the review of papers published between 2015 and 2024, we have classified the presented approaches and offered practical taxonomies, which provide researchers with critical perspectives and highlight potential areas for future exploration and research. Our investigation shows how combining blockchain's decentralized framework with DRL can address privacy and security issues, improve mobile transmission efficiency, and guarantee robust, privacy-preserving IoT systems. Additionally, we explore blockchain integration for DRL and outline the notable applications of DRL technology. By addressing the challenges of machine learning and blockchain integration, this study proposes novel perspectives for researchers and serves as a foundational exploration from an interdisciplinary standpoint.

# 1. Introduction

In the modern world, the Internet and the Internet of Things have changed how we interact and work together. As a result of IoT devices' variety and resource constraints, security has emerged as one of its most significant challenges. This enables ubiquitous computing services to be provided to end users through the interconnection of various devices monitored and controlled via the Internet. IoT is expected to undergo constant revolution over the next few years as a result of numerous constraints, including heterogeneity of devices, resource constraints, power storage, security, and data management. Several attacks are aimed at Internet and IoT-based security, including spoofing and single points of failure. Many of these attacks rely on centralized trusted authorities, making it vulnerable to many attacks. Blockchain will most likely meet the security needs of the Internet and IoT since it provides security, anonymity, and integrity without requiring third parties [1]. In this way, IoT and machine learning will form the basis of Industry 4.0. By facilitating data acquisition within an application context, the IoT is helping digitize many applications and automate business processes [2-5] through deep learning (DL) and deep reinforcement learning (DRL) algorithms. Around six times more data will be generated by IoT devices than their predecessors. Security and privacy protection are of the utmost importance when it comes to IoT applications that involve sensitive data. Additionally, devices are reluctant to share their data for training purposes in open environments like the Internet out of fear of being tracked. Various attackers may manipulate data shared by geographically dispersed individuals, resulting in inaccurate results.

Moreover, IoT systems must be autonomous so they can make decisions based on context and learn from the data they collect. It can help recognize patterns, evaluate, process, and make intelligent judgments in such a setting using machine learning (ML). Despite this, most accessible machine learning algorithms rely on a centralized framework, which may lead to security breaches. Data manipulation, forged authentication, and privacy protection are all challenges of centralized power. The Deep Reinforcement Learning algorithm also depends on data integrity to provide reliable results. Certain occurrences may be falsely identified by Deep Reinforcement Learning algorithms even if there is a modest security flaw.

Furthermore, for many security applications, Deep Reinforcement Learning relies on a trusted third party (TTP) (e.g., a cloud service provider) for calculations, which may compromise privacy. It is, therefore, necessary to develop a Deep Reinforcement Learning algorithm based on a decentralized architecture. Meanwhile, Deep Reinforcement Learning algorithms can be implemented in blockchain and IoT to enhance the security and privacy of blockchain networks. The general organization of the paper is shown in Figure 1.

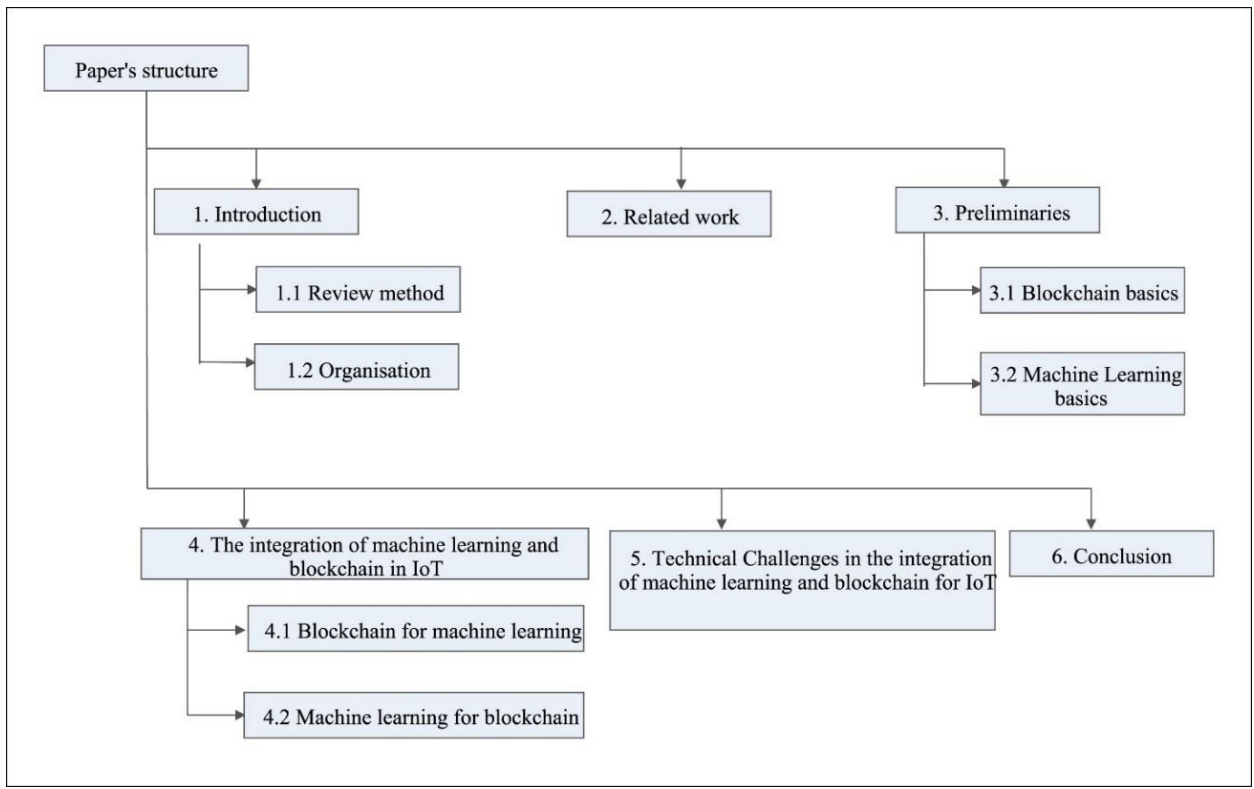

**Fig. 1.** An overview of the survey's structure.

The blockchain network can be used to make predictions or analyze data using machine learning models. In this application, data is collected by multiple sources, including sensors, smart devices, IoT devices, and the blockchain, and the machine learning model can be applied to the data to provide real-time data analytics or predictions. As an integral part of the application, the machine learning model works as a central part of the application. Because the data in the network of blockchain will not contain missing values, duplicates, or noise, the machine learning models will be more accurate since the data in the network will not have those things. The following image illustrates BT-based machine learning adaptation architecture.

The purpose of this article is to describe and illustrate how to identify and cluster IoT application systems using DRL approaches in order to enhance their performance. DRL was also implemented into blockchain-enabled IoT applications as part of this research. The purpose of our paper is to discuss the integration of machine learning and blockchain technology into IoT systems. The methodology of the survey was described in Section 1. The following sections discuss relevant surveys related to blockchain integration and machine learning. In addition, we highlight the differences between this survey and previous comparable surveys. Machine learning and blockchain are discussed in Section 3. This article provides an overview of the language, core concepts, and recent advances in the field. Section 4 presents three types of advanced literature on machine learning and blockchain: reinforcement learning, Internet of Things, and cybersecurity. This section compares the literature on blockchain and machine learning. Future research should address concerns related to machine learning and blockchain technology.

## 2. Related work

Systematic Literature Review (SLR) results do not support the use of Blockchain smart contracts for protecting the Internet and IoT [6-9]. Researchers reviewed the literature to determine whether Blockchain can be used for cybersecurity purposes. Blockchain is being used for Internet and IoT security in more than half of the literature reviewed. Future projects should evaluate the potential for developing novel cybersecurity solutions using Ethereum and other Blockchain platforms. According to Sharma et al. [8], Blockchain-based systems can also be used to deliver security services in areas such as confidentiality, integrity assurance, authentication, privacy, access control, and data provenance. To distinguish between study themes, various studies related to our issue will also be discussed. Security concerns and potential solutions for the Internet of Things were categorized using a tiered taxonomy and classification system. Additionally, they examined and discussed Blockchain-based security solutions to safeguard IoT. A review of Blockchain-based applications was conducted by Billah et al. [10]. Blockchain technology and the Internet of Things were first modeled and examined. Based on the existing work on enabling Blockchain technologies, physical solutions (PSs) were defined to identify the limitations and critical areas for the implementation of Blockchain technology in large-scale, distributed settings. In order to provide IoT security assistance, three tiers of functional requirements (FRs) were identified. As a result of previous research, they categorized physical security solutions into ten categories. Most of these solutions use smart contracts to guarantee a certain level of security. Ogundokun et al. [11] reviewed blockchain-based applications across a range of fields in 2022. From supply chains to IoT, they categorize Blockchain-based applications, focusing on their limitations, which hamper their widespread adoption.

Machine learning algorithms and methods are widely used in a variety of applications, but they have been constrained by the enormous amount of data involved in training. As the number of input parameters used to train a neural network increases, the storage and processing requirements may escalate dramatically. In problem domains where individual devices are resource-constrained, UAVs are ideal for hosting machine-learning algorithms that require large amounts of processing power. The federation notion in machine learning is illustrated realistically by Ogundokun et al. [12]. Mobile devices are used to construct machine learning models where the data exists, and TensorFlow is used to train deep neural networks from the model's individual data in the cloud [13]. Implementation challenges such as localization (time zone), device availability, and limited computing resources have been recognized and solved by the authors. A server with the updated model communicates with the devices. After model training, the device updates the model and sends it back to the server with the updated model. Federated learning activities use a protocol that is designed to be resilient to unreliable settings in terms of communication frequency, device involvement, and device selection. An analytics method allows the server to gain insights about bottlenecks and other restrictions occurring at the device's end without accessing its data. A variety of techniques were used by Chen et al. [14] to optimize machine learning. Part of the data reflected individual user behavior in the optimization of a variety of mobile devices. Their focus is optimizing and enhancing machine learning models using federated learning. The authors investigate the challenges in achieving optimal performance when using known algorithms. Distributed Approximate Newton (DANE) [15] and Stochastic Variation Reduced Gradient (SVRG) [16] are two distributed machine learning techniques.

A typical machine learning model performs poorly in this context due to its intrinsic sequential nature. This model is updated using the parameters learned by the nodes through a communication round. There are rapid iterations in traditional algorithms. Based on experimental data, they demonstrate that it is feasible to build an algorithm that performs well in this environment despite the limitations of network capacity and obstacles such as sluggish convergence. According to Chen et al. [16], distributed machine learning had one of its most significant contributions. Specifically, they contribute mechanisms for sharing training data and workloads among several worker nodes. Additionally, the proposed architecture enables the global sharing of model parameters in the form of sparse vectors between server nodes. There are billions of samples being considered here, with potentially very long feature lengths. Workers train on subsets of the whole data set, while parameter servers store copies of the most recent parameters. Latent Dirichlet Allocation and Sparse Logistic Regression are the authors' machine learning models. Thanks to a bulk communication server and message compression during parameter transmission, the system operated effectively because of reduced communication costs. Xu [17] proposes that each node executes its own training algorithm to gather datasets from distant nodes. Based on variables such as which model to use, what features to pick, and how long a given configuration will take, this system is designed to assist data pre-processing activities such as feature selection, engineering, and optimization. This concept also inspired the approach proposed in this study. To create a machine learning model that is ready for deployment, we offer this feature in our proposed framework to simulate human behavior during data preprocessing, training, and use. To enable privacy-aware federated learning, we also use blockchains to allow the trusted sharing of models and their performance indicators. In [54], Luong et al. studied DRL applications in communications and networking, focusing primarily on networking and communications issues. The use of DRL for autonomous IoT was surveyed by Lei et al. [55]. However, despite the usefulness of these studies, there are still some emerging challenges that they have not addressed. In the case of blockchain-powered IoT applications, for example, strict trust may be required. A number of optimization algorithms for network scheduling and duty cycling are then investigated by Bai et al. [56] and Wang et al. [57].

Blockchain has recently been used to maximize the utility of machine learning algorithms in a number of noteworthy contributions. Using smart contracts to test and verify machine learning models is illustrated by Liu et al. [18]. Due to the Ethereum-based blockchain employed, users can specify a reward amount in Ethereum tokens once they upload a dataset to the blockchain. After the machine learning task is completed by the human-driven agent, the machine learning model is presented to the smart contract that certifies the answer. The contestant with the best model is awarded the prize once all competitors have successfully submitted their ideas. Machine learning preprocessing and model-fitting are automated in our case to simulate this behavior on the nodes. By submitting datasets on the blockchain and paying the top model submissions based on their performance, the authors have developed a prototype for a future marketplace where enterprises can obtain the best available machine-learning skills. Smart contracts are used to control trust and equity as well as model validation and compensation. The authors have also identified the dangers associated with cheating by model submitters and organizers and implemented smart contracts to deal with them. As an additional feature of this study, the machine learning technique has been implemented with Solidity, which is a computer language. It was necessary for the authors to incorporate several speed and memory-saving strategies, given the language's limitations.

**Table1**. Key findings and themes of primary studies.

| Ref. | Findings | Platform for blockchains | Domain of primary application | Services related to security |
|---|---|---|---|---|
| [58] | An IoT data management architecture based on Blockchain, fog computing, and the cloud. Smart contracts are used to monitor access and manage resources in the proposed architecture. | Ethereum | IoT | Access Control & Secure Data Management |
| [59] | Utilizing the layered architecture of Blockchain smart contracts to meet the functional and non-functional needs of heterogeneous IoT. | Ethereum | IoT | Provenance of data |
| [60] | Data provenance and data integrity can be enforced with Ethereum Smart Contracts with Physical Unclonable Functions (PUFs). | Ethereum | IoT | Integrity & Provenance of Data |
| [61] | Using Ethereum smart contracts for safe transmission of digital material (updates for IoT devices). | Ethereum | IoT | Fault tolerance & device integrity |

| Ref. | Findings | Platform for blockchains | Domain of primary application | Services related to security |
|---|---|---|---|---|
| [62] | IoT access control using Smart Contracts, a Blockchain-based technique. | Custom | IoT | Access Control |
| [63] | An experiment in enforcing GDPR regulations with blockchain-based smart contracts for IoT data production. | Ethereum | IoT | Audit Trail and Data Protection |
| [64] | Smart contracts and PUF answers are used to detect counterfeit IoT nodes. | Ethereum | IoT | Malicious Node Detection |
| [65] | By using static resource allocation, IoT devices can be prevented from gaining access to the server and DDoS attacks can be prevented. | Ethereum | IoT | DDoS Protection |
| [66] | Utilizing smart contracts to safeguard IoT firmware upgrades. | Hyperledger Fabric | IoT | Device Integrity |
| [67] | Smart Contracts and Open Provenance models enable a Blockchain-based platform to gather, verify, and validate provenance information securely. | Ethereum | IoT | Data Provenance & Access Control |
| [68] | Using blockchain technology to provide fair non-repudiatory services to the IIoT. On the blockchain, smart contracts act as service publishers, evidence recorders, and conflict resolution mechanisms. | Ethereum | IoT | Indefatigability |
| [69] | It ensures Device Integrity, Authentication, and Non-Repudiation through a blockchain-based IoT control protocol. | Ethereum | IoT | Integrity, authenticity, and non-repudiation. |
| [70] | An Internet-of-Things firmware upgrade security method based on Grammar and Compiler. | Hyperledger Fabric | IoT | Integrity of data and devices |
| [71] | Internet of Things access control method based on smart contracts. | Ethereum | IoT | Controlled access |
| [72] | Blockchain-based key-value databases and IoT access control system with fine-grained security, scalability, and safety. | EOS | IoT | Controlled access |
| [73] | By combining blockchain-based trading platforms with proxy reencryption without pairing, sensor data transmission can be ensured to be as secure as possible. | Ethereum | IoT | Protection of data |
| [74] | WSNs' quadrilateral measurement localization approach and blockchain smart contracts are used to identify malicious nodes in WSNs. | Ethereum | IoT | Detection of malicious nodes |
| [75] | For IoT-enabled PV systems based on smart contracts and blockchain technology, an access control and data integrity framework is developed. | Hyperledger Fabric | IoT | Detection of malicious nodes |

| Ref. | Findings | Platform for blockchains | Domain of primary application | Services related to security |
|---|---|---|---|---|
| [76] | It consists of a decentralized key-store containing the public keys of all devices and a generic protocol for distributing PSKs based on the Ethereum blockchain. | Ethereum | Internet & IoT | Authentication & Integrity of Data |
| [77] | Secure data sharing management can be achieved through Blockchain-based Smart Contracts. | Ethereum | IoT | Protection of data |
| [78] | A technique based on Ethereum and PKI to authenticate and distribute PSKs for securing IoT communication channels. | Ethereum | IoT | Verification |
| [79] | As part of a secure mutual authentication system, blockchain, group signatures, and message authentication codes are used. | Ethereum | IoT | Verification |

An energy and resource management system based on blockchains is presented by Muneeb et al. [19]. By using machine learning, smart contracts reduce energy costs in the data center progressively. Smart contracts that control data center resource utilization use the output of the machine learning model. Blockchains and machine learning have been used in several cyber security initiatives. Zhang et al. [20] built a trustworthy, self-adjusting solution with blockchains and machine learning for IoT access management that can be continuously updated through reinforcement learning, for example. Using machine learning, Ghazal et al. [21] identified anomalous behavior in blockchain networks such as collusion that can lead to majority attacks. The assault is countered by using algorithmic game theory and supervised machine learning. Ethereum smart contracts share locally learned gradients, according to Rodríguez-Rodríguez et al. [22]. They separate data storage from data processing nodes to achieve privacy-aware data processing.

To facilitate the development of innovative solutions in this sector, machine learning and blockchain technology are being integrated to facilitate the development of autonomous cars and unmanned aerial vehicles (UAVs). Machine learning plays a key role in the efficient and effective detection of unmanned aerial vehicles (UAVs) in this field. Several machine learning approaches have been developed for the identification and classification of UAVs, including [23-26]. Current attempts to simplify the development of trustworthy applications for UAVs have been highlighted by [27-30]. Unmanned aerial vehicles (UAVs) can support trustworthy distributed machine learning systems by using blockchain technology. To maximize the benefits of a UAV-based system's distributed architecture, the proposed framework utilizes a federated machine learning approach while reducing the performance overheads caused by conventional centralized approaches. A further benefit of the proposed system is that it employs blockchain technology to facilitate the sharing of machine learning models and related metrics in a trustworthy manner. An intrusion detection scenario is used to evaluate the effectiveness of the proposed system and uncover potential benefits as well as unresolved issues (See Table 1).

## 3. Basics of blockchain technology

In 2008, Satoshi Nakamoto released a fictional currency called Bitcoin that was based on blockchain technology, which is widely regarded as a decentralized, transparent, and trustworthy ledger [10]. Each transaction on the blockchain is packaged into a block, which contains a certain number of transactions. Distributed blockchain ledgers are created using all verified blocks. Using the cryptographic hash code of a block, a block in the distributed ledger is connected to a previously authorized block. A variety of uses outside of digital currencies have already been explored with this developing technology. In a P2P network, members can check the behavior of other participants, create new blockchain-recorded transactions, and validate them. As a result of this architecture, blockchain operations are robust and efficient, and the risks of single points of failure are reduced. Network administrators have yet to govern the blockchain ledger, which is accessible to all participants [11-15]. In order to achieve consensus among network nodes, rigorous rules must be enforced, and strict rules must be enforced. This is known as the consensus method. In a blockchain network, consensus is the process by which all nodes sync the distributed ledger. the Bitcoin blockchain is illustrated in Figure 2.

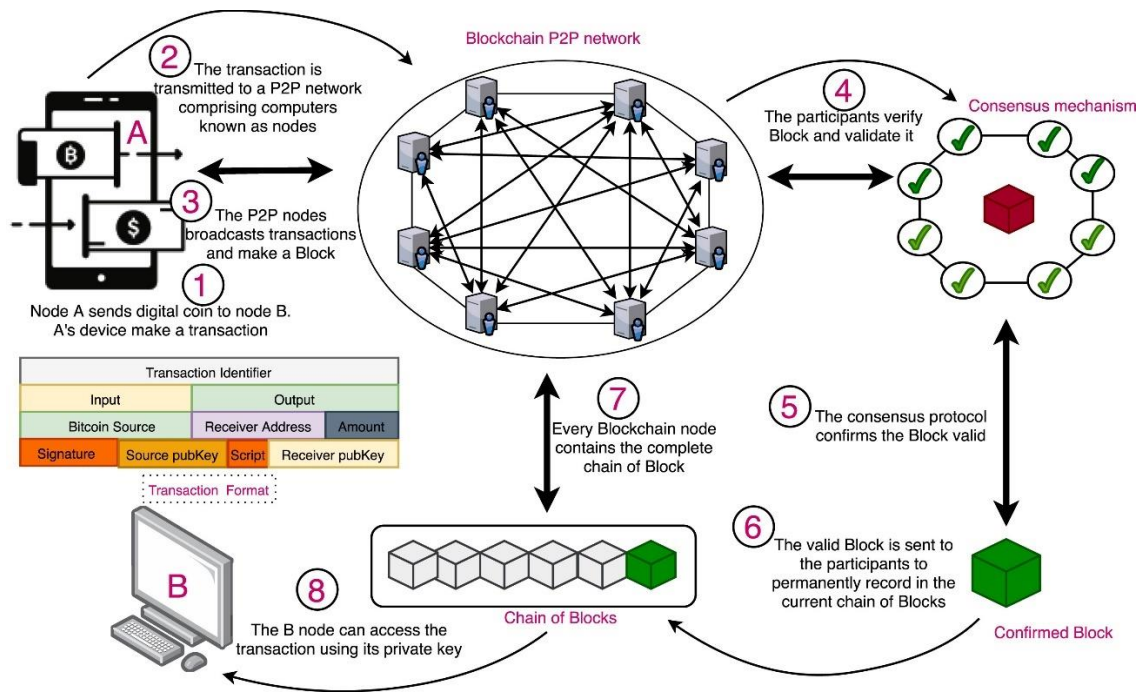

**Fig. 2.** The Bitcoin blockchain's basic operation (adapted from Ref. [35]). Peer-to-peer: peer-to-peer.

## 2.1. IoT and blockchain (BCIoT) in Smart City

Embedded sophisticated computers, sensors, and actuators transmit data to a centralized server, usually a Cloud server, in order to enable the IoT to connect people, things, and commodities. IoT analytics tools are used to improve corporate operations and develop new services based on IoT data. While the IoT ecosystem has a lot of potential, security and privacy are important concerns that have hindered its widespread adoption [16, 17]. It is common for IoT networks to be vulnerable to security flaws such as DDoS attacks, ransomware, and malicious attacks. In DDoS attacks, several hacked computer systems bombard a target with many simultaneous data requests, resulting in a denial of service for users. The current centralized systems may encounter bottlenecks as the number of devices connecting to an IoT network grows. DTL blockchain potentially overcomes the security, privacy, and scalability issues associated with IoT by resolving these IoT issues. A blockchain's distributed ledger can't be tampered with, so participants don't have to trust each other. Internet of Things has a wide range of applications, including smart cities, smart infrastructure, smart grids, smart transportation, and smart homes. A new blockchain domain, BCIoT, has emerged with the implementation of blockchains in the IoT domain. Data produced by IoT devices is not controlled by one organization under the BCIoT paradigm [18, 19]. Furthermore, parties can review previous transactions with blockchain technology. As a result, data leaks are identified and repaired quickly. It is becoming increasingly important to ensure the integrity of IoT source code since it is held by internet third parties and telecommunications companies. The usefulness of blockchain in IoT networks is affected by several variables [20, 21]:

(1) In IoT applications that require a decentralized P2P ecosystem, blockchain can solve privacy and security issues.

(2) IoT applications requiring payment processing without third-party control may benefit from blockchain technology [22].

(3) Blockchain can be an effective solution if IoT applications require logs and traceability of sequential transactions.

IoT devices and blockchain ledgers work well together, but there are some key obstacles to overcome.

- IoT sensors produce a vast amount of data that must be handled on-chain in order to integrate IoT and blockchain. Transactions on the blockchain may also be processed at lower speeds or with a higher latency.
- Keeping network privacy and transaction confidentiality is also key: on a public blockchain, it is not possible to ensure anonymity for transaction history. The analysis of transaction patterns can be used by attackers to discover the identities of users or devices.

After reviewing the literature, we focused on IoT and blockchain to address the issue mentioned above. An integrated architecture of IoT and blockchain architecture is demonstrated in the following figure (figure 3):

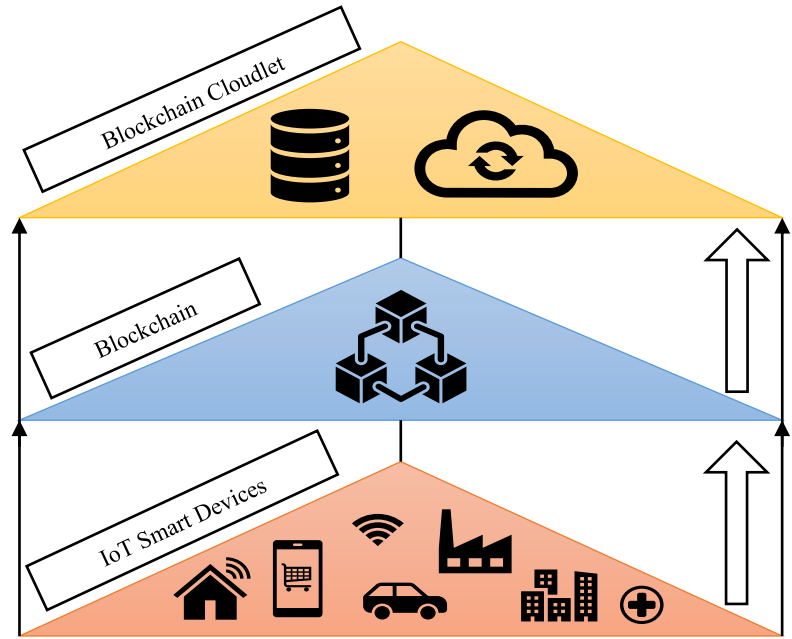

**Fig. 3.** An Integrated IoT/Blockchain Architecture

Awotunde et al.[95] explored privacy and security issues facing smart cities and proposed a hybrid deep learning-enabled blockchain solution to improve security and privacy. Several security and privacy concerns are associated with the rapid adoption of IoT-powered smart cities, including data breaches, unauthorized access, and service disruptions. The authors stressed the importance of robust security in smart cities, as weak systems can be exploited by malicious actors to gain access to networks or sensitive information. Using Singular Value Decomposition (SVD), Autoregressive Integrated Moving Average (ARIMA), and blockchain technology in the Industrial Internet of Things (IIoT), Miao et al.[96] introduce a novel short-term traffic prediction model.

The model is intended to address the challenges of nonlinearity, non-stationarity, and complex structures inherent in traffic data. We used SVD to extract features from historical traffic data, ARIMA to model temporal dependencies, and blockchain technology to ensure the integrity and security of the data. It was found that the proposed model outperformed traditional statistical and deep learning approaches in terms of accuracy and real-time performance, showing its effectiveness in addressing traffic prediction challenges in smart cities. Huda et al.[97] conducted a comprehensive study on the relationship between technology, automation, and sustainability in the transformation of smart homes into sustainable smart cities. This study concluded that integrating human intelligence with technological advancement leads to synergy and collaborative innovation, thus fostering urbanization. In addition to enhancing comfort, convenience, and living standards, smart home automation is an IoT-based subset. In the transition to sustainable smart cities, technology and automation play a significant role in achieving sustainability. Bhatia et al.[98] reviewed the role of machine learning in healthcare. According to the study, the integration of blockchain and artificial intelligence could revolutionize data management across a variety of industries, including healthcare, finance, and supply chain. Data can be securely stored and shared using the authors' innovative framework, which leverages the strengths of both technologies.

A secure platform for digital governance interoperability and data exchange was developed by Malik et al.[99] utilizing blockchain technology and deep learning frameworks. A seamless information sharing platform between agencies and departments enhances trust in digital governance by ensuring privacy, security, and reliability. Based on deep reinforcement learning (DRL) and secured blockchain technology, Li et al.[100] presented a novel approach to energy management in smart cities. Increasing urban energy demand requires intelligent solutions to optimize energy consumption while maintaining data security and integrity. The authors propose integrating DRL algorithms with blockchain technology in order to achieve optimal energy management, as well as address security concerns. In order to enhance resource utilization efficiency, their DRL framework dynamically adapts to changing energy demands and environmental factors, continuously learning and optimizing energy allocation strategies. A secure web service model based on starling murmuration was proposed by Sheeba et al.[101] for smart city applications utilizing deep belief networks (DBN). A time-consuming and expensive manual analysis of Web Service Description Language (WSDL) documents was the objective of the study. A structural self-organized DBN (SSODBN) architecture was used in combination with a dropout strategy to minimize interrelationships between feature detectors. It has been applied to a variety of fields, including education, smart electricity, intelligent road networks, health and social care, sports, water, and gas distribution for the classification of real-time web services. Based on the results, the SSODBN-based model accurately classified web services across different domains, providing a novel solution to the limitations of manual WSDL analysis. Based on deep reinforcement learning (DRL) and secured blockchain technology, Li et al.[6] presented a novel approach to energy management in smart cities. Increasing urban energy demand requires intelligent solutions to optimize energy consumption while maintaining data security and integrity. The authors propose integrating DRL algorithms with blockchain technology in order to achieve optimal energy management, as well as address security concerns. In order to enhance resource utilization efficiency, their DRL framework dynamically adapts to changing energy demands and environmental factors, continuously learning and optimizing energy allocation strategies.

**Table 2**: A Summary of Literature in IoT and blockchain in Smart City

| Authors | Year | Objective | Approach | Conclusion |
|---|---|---|---|---|
| Almuqren et al.[105] | 2023 | Blockchain and deep learning can be used to improve the security of smart home networks | A smart home network based on blockchain technology that employs gradient-based optimization combined with hybrid deep learning models | Enhanced security by ensuring data confidentiality, authenticity, and availability while mitigating the risk of cyber-attacks |
| Chen et al.[106] | 2024 | Utilize blockchain, IoT, and edge computing to optimize urban traffic management | Smart Traffic Management System (STMS) based on Twin Delayed Deep Deterministic Policy Gradient (TD3) reinforcement learning integrated with blockchain | Boosted urban traffic management through secure and transparent data sharing, optimizing traffic flow, and reducing congestion |
| Jin et al.[107] | 2023 | Optimize task offloading in blockchain-enabled smart cities | Deep reinforcement learning framed as a Markov decision process for task offloading | Optimizing task offloading has improved Quality of Service (QoS) and increased profits |
| Kumar et al.[108] | 2024 | With advanced encryption and attack detection, smart city data can be securely stored in the blockchain | Combining advanced encryption techniques with machine learning algorithms to detect potential attacks on blockchain-based smart city data | Identified and prevented various attacks, emphasizing the importance of strong security measures |
| Kumar et al.[109] | 2024 | The use of blockchain and explainable artificial intelligence to secure | Combining blockchain-based authentication with | Reduced computation and communication costs by providing efficient |

| | | consumer IoT applications in smart cities | artificial intelligence-based intrusion detection | authentication and key agreement mechanisms |
|---|---|---|---|---|
| Liu et al.[110] | 2023 | Assess the use of blockchain, IoT, and edge computing in healthcare services within smart cities | A comprehensive evaluation of these technologies is required | IoT plays a crucial role in healthcare, edge computing allows for low-cost remote access, and blockchain enables secure data assessment |
| Hsieh & Chen.[111] | 2023 | Analyze the development of knowledge in the area of the internet of vehicles (IoV) | Data mining, citation networks, cluster analysis, and main path analysis of studies from Web of Science | Identified key research topics and knowledge development paths in IoV, such as media access control and V2V channels |
| Omar et al.[112] | 2024 | Implement a blockchain-based system for managing road traffic | Blockchain-enabled IoT sensors, cameras, and deep learning models for detecting and logging traffic violations | Detection and logging of traffic violations using blockchain-enabled IoT sensors, cameras, and deep learning models |
| Ali et al.[113] | 2023 | Blockchain and deep learning can enhance the scalability and security of healthcare systems | Integrating deep learning techniques with blockchain technology | Maintaining data integrity, privacy, and interoperability without relying on centralized authorities |
| Ramalingam et al.[114] | 2023 | Integrate blockchain technology with computer vision for a variety of applications | Literature review of blockchain-based computer vision systems | Supply chain, healthcare, smart cities, and defense sectors could benefit from improved transparency, accountability, and security |
| Mishra & Chaurasiya.[115] | 2023 | The use of blockchain and IoT to secure smart city infrastructure | An IoT framework based on blockchain technology that uses Adaptive Data Cleaning, denoising autoencoder methods, and deep learning algorithms optimized by Dingo | Enhanced security and efficiency in the development of smart cities |
| Jadav et al.[116] | 2023 | Using blockchain and garlic routing for secure data exchange in MTC-based IIoT | For secure data exchange in IIoT beyond 5G, GRADE integrates blockchain and garlic routing | Managed the secure sharing of data in the IIoT, addressing security challenges in MTC applications |
| Huynh-The et al.[117] | 2023 | Enhance immersive experiences and human-like | A survey of AI applications in virtual environments | Improved realism and interactivity of virtual agents, allowing for natural |

| | | interactions in the metaverse through the use of artificial intelligence | | conversation and emotional connections with users |
|---|---|---|---|---|
| Aldhyani et al.[118] | 2023 | In smart cities, develop a secure IoMT framework for breast cancer detection | Data collection and machine learning-based diagnosis using AI, computational intelligence, and the Internet of Things | Achieved high accuracy in the detection of breast cancer, ensuring secure communication and minimizing environmental impact |
| Ranjith & Malagi.[119] | 2023 | Data acquisition techniques for smart transportation systems should be improved | An overview of modern approaches to data acquisition, including hierarchical deep reinforcement learning, energy-efficient UAVs, and swarm intelligence | A focus was placed on the efficient processing of large datasets in smart transportation applications |
| Sirohi et al.[120] | 2023 | Implement federated learning in 6G-enabled secure communication systems | An analysis of federated learning applications | A study of federated learning applications |
| Alsolami et al.[121] | 2023 | Using deep reinforcement learning to optimize energy trading in smart grids | Optimizing P2P trading in smart grids using deep reinforcement learning | Enhanced energy trading and demand response, resulting in a reduction in household electricity costs |
| Mishra & Chaurasiya[122] | 2024 | Blockchain and IoT for secure transactions in smart cities | Data preprocessing using hybrid deep learning LSTM-SVM algorithm with min-max normalization and weighted average filter | Detection and prevention of cyber threats in smart city transactions |

A multi-objective reinforcement federated learning system for transport data in smart cities was examined by Mohammed et al. [102]. Their goal was to address the cybersecurity challenges associated with heterogeneous cloud computing-based servers that provide communication and computation services. In this study, reinforcement federated learning was integrated with blockchain technology in order to enhance the security and efficiency of IoT-enabled transportation systems. As part of the solution, a decentralized architecture was used to facilitate secure data sharing among multiple stakeholders while maintaining the integrity of transport-related information.

A hybrid framework was employed by Shankar & Maple.[103] to investigate the security challenges in IoT-enabled smart city infrastructure. As a result of the diversity and novelty of IoT applications, the authors acknowledged that it is impossible to develop a single, universally applicable design for all IoT applications. In order to identify new societal benefits, more research is required. This study highlighted the importance of ethics and technology in ensuring the safety and security of individuals in smart cities, and the need for multiple recommendation designs to coexist in the IoT landscape. According to Devarajan et al.[104], a blockchain-enabled secure federated learning system model for traffic flow prediction in urban computing can overcome fog computing limitations in vehicular networks. Results showed that the proposed BSFLVN system achieved better performance and security than existing methods, demonstrating the effectiveness of integrating blockchain technology with federated learning to address privacy concerns and increase efficiency. In order to achieve smart cities, fog computing-based systems must balance security and efficiency. A comprehensive summary of current literature regarding IoT and blockchain applications in smart cities is provided in Table 2.

## 2.2. Blockchain and Cloud of Things (BCCoT)

As digitalization improves, health institutions and patients exchange a growing number of electronic medical records (EMRs), enabling data collection and enhancing the quality of treatment for patients. Using cloud computing, patients may remotely access their Electronic Health Records (EHR) on their mobile devices, and EHRs can be remotely processed on Cloud servers. With IoT and Cloud computing, therapy may be supplied on demand, medical costs can be decreased, and care quality can be enhanced [21, 22]. Despite this, cloud-based storage, Cloud suppliers, and users are not fully trusted to exchange data in Cloud contexts. In addition to deteriorating medical services and networks, this causes serious data leakage issues. Due to the immutability, stability, and reliability of blockchain technology, exchanging health information in cloud ecosystems presents obstacles [23]. Blockchain and Cloud can secure data sharing and are significant contributors to controlling user access and data sharing in Cloud IoT-enabled healthcare networks. Using smart contracts on the blockchain, unsafe healthcare environments may be regulated and authorized automatically, assuring their security. To secure a high level of data privacy and security, blockchain models promote participation by patients and healthcare organizations. When blockchain is used with cloud computing, eHealth cloud storage services are substantially more secure. Under blockchain management, cloud storage functions as peers inside the P2P network. Researchers have proposed encrypting and storing health data in normal cloud storage while storing the hash code created from metadata on the blockchain. This will allow cloud data traceability and rapid change detection. Healthcare services can be delivered more efficiently, dependably, and productively with blockchain technology. Using blockchain technology to monitor patient health, diagnose patients, and evaluate therapeutic procedures could revolutionize the clinical sector. Thus, blockchain models will contribute to the transformation of healthcare delivery into a more secure and efficient procedure [123-125].

Smart city applications can benefit from blockchain's superior security features. In addition to providing sophisticated computational capabilities, cloud computing allows individuals to provide services in real-time based on huge data streams from IoT applications. Smart cities can benefit from blockchain's high-security characteristics when managing operations [126-129]. Smart city designs can address performance and security issues with blockchain and cloud computing. Home monitoring, home management, and device access control are some of the smart home services provided by blockchain systems. Data storage and processing between IoT devices, homeowners, and external users may be more scalable and efficient using Blockchain and distributed Cloud computing. In the present IoT systems, the massive amount of data transmitted from several devices causes a bottleneck due to the limited power and storage capacities of IoT devices [127, 28]. Today, most data storage and processing are done in a central database. There are several disadvantages to a centralized repository [130]:

(1) Consumers would not be able to access services during the outage since a single server handles all user requests.

(2) Businesses that manage centralized storage mediums may disclose unencrypted data to unauthorized individuals [29], placing the data owner's privacy at risk.

(3) It is possible for the database to be modified from the server without the data owner's knowledge or consent [7, 30].

Cloud computing, however, can provide IoT services on-demand, dependable, and secure due to its nearly limitless storage and processing capabilities. Blockchain, IoT, and cloud computing create a new paradigm, BCCoT, which secures the operation of apps. Cloud-based resources are extremely beneficial to IoT frameworks. Real-world applications may also become more prominent due to Cloud's integration with IoT ecosystems. A Cloud of Things system could also lead to a more efficient, effective, and high-quality IoT system with little management effort [131]. There are several IoT processes that can be supported by cloud-based analytical tools, including the processing of historical data, storing and analyzing information, and analyzing statistical data. Cloud data management is used to enhance IoT services and meet consumer needs. The Internet of Things (IoT) has many key characteristics, such as ubiquitous connectivity, on-demand assistance, high processing capacity, and scalability [131,132]. To combine blockchain, the Internet of Things, and Cloud of Things technologies, academics have designed a variety of frameworks.

### *2-3-Nexus of machine learning, blockchain, and IIoT*

Blockchain systems can be classified as public [114], consortium [15], or private [16]. Their distinctions are illustrated in the following table. All data stored on a public blockchain is accessible to the public. The data on public blockchains is open to all, so anyone can read, write, and see it. As a result, blockchains can be utilized in areas such as healthcare. Compared to public blockchains, private blockchains are often used inside organizations and are not publicly accessible [133, 134]. The public cannot access private blockchains. Blockchains that are administered by companies are called private blockchains. In a private blockchain environment, records can be kept securely, audited, and tampered with. Since public blockchains are less regulated than private ones, validation and storing transactions take less time. Transaction speeds on private blockchains may rival those on traditional databases [134-139]. Due to the normal safe network environment, a limited number of nodes may be trusted. A private blockchain can be used for manufacturing, for example. Blockchain consortiums are hybrids of public and private blockchains, although they are more closely related to private distributed ledgers. Businesses using "semi-private" systems have restricted access to them. It is both more expensive and time-consuming to administer traditional Internet of Things applications due to centralized administration. IIoT challenges will be alleviated by the deployment of blockchain technology because of the rapid proliferation of IoT devices. Each node in a blockchain network functions as a peer node, so there are no additional protocols required for peer-to-peer interactions [133]. By eliminating the requirement to convey transaction information to third parties, a blockchain system allows peer-to-peer data sharing. By storing redundant data on the blockchain, multiple nodes (peers) can preserve data integrity and provide flexibility for IoT systems. It may be possible to make an IoT system more adaptable and interoperable with a variety of devices and data formats by using blockchain technology [140]. As a result, the IIoT is equipped with vast amounts of data that are tailored to certain business needs, like privacy and transparency. In order to build and deploy deep learning and reinforcement learning methods, as well as to sustain and automate business processes, data availability is imperative [134-139].

**Table 3**. A Side-by-Side Comparison of Blockchain Types.

| Cell with no contents | Blockchain for public use | Blockchain consortium | Blockchain technology for privacy |
|---|---|---|---|
| Assistant | Anybody | Consortium members | Group members |
| recorder | A group of participants | Discussions | Individualization |
| Rewards | Mandatory | optional | Optimal |
| Centralization | Centralization | multicenter model | multi-center |
| Disclosure of data | Generally, | Limitations | Confidential |
| Relationship-based on trust | Insufficient | average | Having strength |
| Efficiency of consensus | Insufficient | Medium-sized | High-quality |
| Sectors of industry | Medical care | The manufacturing industry, energy, water, and finance industries ||

A device connected to the Internet of Things can be a peer device in the infrastructure (a miner) or a user device supported by a miner (e.g., a medical device, a mechanical arm, a warehouse robot, a monitoring sensor, an autonomous car). As these devices are primarily designed for ease of manufacture and/or application rather than for storing and calculating, their storage and computation capacity is often limited [141-145]. The capabilities of some nodes should have been designed to facilitate production and application rather than establishing consensus and storing data that is not part of their original design. To reduce computational overhead and maximize storage resources, blockchain implementations should use a suitable consensus technique in the IIoT to reduce computational overhead. Blockchain communication networks can be constructed in a variety of ways [140]. In a smart factory, for example, wireless networking may be a good choice for devices and equipment [18]. Device-to-device (D2D) interactions may enhance the resilience of the system and allow for better utilization of cellular and WiFi networks. [119-124]. Thus, networking is a topic worth investigating. Several computer and control systems have single points of failure, which must be addressed before IIoT can be widely used. (Table 3).

## *2-3-1- Supervised Learning*

It involves labeling both input data as well as targeted output data for classification in supervised learning. Future data processing will be based on this learning base. There are a number of key algorithms for supervised learning, including:

1- The k-nearest neighbor (k-NN) approach involves classifying data samples according to their neighbor's labels. The average measurements of neighboring IoT devices within a specific range are typically calculated using simple methods (e.g., the Euclidean distance between the devices). In large data sets, it may not be accurate due to the simplicity of the computation algorithm. k-NNs are used in IoT to detect faults and aggregate data.

2- To define decision boundaries in SVM approaches, decision planes are used. As a result of its high accuracy, SVM supervised learning is typically used for localization problems in IoT and WSN in order to detect malicious behaviors.

3- Artificial neural network (ANN): A neural network consists of layers of artificial neurons that are interconnected to mimic the function of biological neurons. As a result of these artificial neurons, different sets of input data are mapped to a set of output data. Although artificial neurons provide solutions to non-linear and complex problems, they are computationally intensive. IoT localization is improved by artificial neural networks.

4- A relatively small number of samples is used for training Bayesian inference, unlike most machine learning algorithms. Adapting probability distributions and avoiding overfitting are the hallmarks of Bayesian learning. They are effective at learning uncertain perceptions. It is necessary, however, for them to be familiar with the environment beforehand. Detecting faults, selecting cluster heads, and localizing clusters can be accomplished with Bayesian interfaces.

5- These decision-support tools use DT (Decision Tree) models to make decisions or classify items. If-then conditions are used to create DTs. A random forest algorithm (RF) is used to improve DT accuracy. The RF method consists of constructing multiple classifiers as part of an ensemble decision tree. Decision trees are the basis of each classifier.

## 3. Reinforcement learning

A decision-making agent using RL actively interacts with its uncertainty during the decision-making process. Dynamic programming extension and MDP are also known as MDP. For closed-loop problems, RL is the answer; closed-loop inputs affect subsequent system outputs. There are no known transition functions or reward functions in this method, unlike the MDP. A RL system consists of the following components:

- In RL, the primary component is an agent that seeks goals in an uncertain environment and interacts with it.
- The environment is everything with which the agent interacts. The environment has certain features and attributes. In each state, the agent senses and evaluates these attributes. Agents use this information to choose actions that could influence outcomes.
- The state of the environment has a direct relationship with actions that can be taken based on it. Searches, functions, and lookup matrices are all options.

- A reward signal defines the goal of a problem. Depending on the agent's current state and action, the environment sends the agent a single value every time step. The reward is this value. It can be positive or negative. This reward is the sole objective of the agent.
- State functions are defined as value functions. Using this state as a starting point and accumulating rewards over time, the agent is said to have collected a value of the state. In order to estimate the values of the agents, iterations are performed over the sequences of observations. Reinforcement learning architecture is demonstrated in the following figure (Fig. 4).

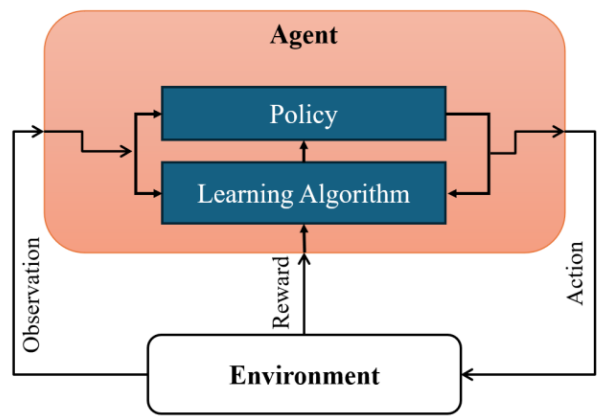

**Fig. 4:** Architecture of Reinforcement Learning

Based on this concept, RL methods can be classified as model-based or model-free. There is no planning or information about previous actions with model-free methods. Methods such as these are based on trial and error [36, 37].

## *3.1. Deep reinforcement learning*

In order to maximize its cumulative benefit, an agent frequently performs high-reward actions from one state to another. In addition, the activity itself is associated with greedy behavior. An agent is investigating whether it chooses an action for transitioning from one state to another about which it has limited or no knowledge. Deep reinforcement learning (DRL) algorithms are employed to understand this behavior further, leveraging neural networks to approximate complex action-value functions. By using these algorithms, agents are able to make informed decisions even in high-dimensional state spaces typical of IoT environments in smart cities.

Simulation of numerous interactions within a digital twin of the indoor model allows the agent to refine its strategy, balancing exploration of unknown states with exploitation of known, high-reward actions. In dynamic environments where new data continuously influence the agent's knowledge base, this balance is imperative for optimizing long-term results. In addition, the integration of collaborative computing enhances the efficiency of DRL in indoor modeling. In the IoT network, agents share insights and strategies, collectively improving policies through a distributed learning process. In addition to accelerating the learning curve, this collaborative approach ensures robustness against isolated errors or suboptimal actions by individual agents. This results in a safer and more efficient indoor environment, which can react quickly to changes and potential threats. Creating responsive and intelligent smart city infrastructure is made possible by this synergy between deep reinforcement learning and collaborative IoT computing. It is possible to give a gift that is far more valuable! Currently, an investigation is being conducted. Alfandi et al. [40] define RL as the balance between utilizing the familiar and exploring the unfamiliar. DL boosts RL's decision-making ability when circumstances require it. According to Mnih et al. [41], DRL played Atari 2600 video games extremely fast and effectively.

### *3.2. Evaluation metrics for reinforcement learning*

A qualitative comparison of RL algorithms in the fields we examined relies on three primary assessment measures.

1. As a performance criterion, the accuracy of authentication is crucial. Metric values used to determine authentication accuracy include.

    a. An invalid user's chances of being accepted are measured by the False Acceptance Rate (FAR).

    b. A Missed Detection Rate (MDR) indicates how likely it is that an invalid user will be missed.

2. Computing performance is influenced by time complexity and time cost. Efficiency is demonstrated by taking less time. There is a reference to greed in an earlier section. As a result of the RL algorithm's computational efficiency, it is considered greedy. An algorithm that selects actions based on their reward is said to be greedy. A learning algorithm may be myopic if it considers an immediate reward. In an ideal situation, greedy action selection would result in the selection of a longer and more advantageous route. It is common for processing efficiency to include storage space efficiency in many applications because most IoT applications operate in resource-constrained environments.

3. Convergence Time - In our previous article, we discussed the role of time as a factor in computer efficiency. The purpose of this paper is to describe the convergence time of a particular RL algorithm. It is likely that an RL agent will conduct thousands, if not millions, of experiments and actions before it reaches optimum learning. It would be ideal if a model could minimize the number of steps needed to reach convergence and interact less with the environment. RL algorithms can be compared using a variety of quantitative metrics.

**4. Cybersecurity, IoT, and reinforcement learning combined**

The security of IoT devices is a primary concern. Many of these devices frequently move across networks, making them vulnerable to danger. Several new laws and regulations have increased the processing overhead for current systems, such as the General Data Protection Regulation (GDPR) and the new environmental legislation [146-150]. In order to be reliable, Machine Learning (ML) malware detection systems must be accurate. RL is an effective and competent machine learning strategy. Our literature review combines reinforcement learning, IoT, and cybersecurity [140-145]. The applications of RL in IDS-IPS, IoT, and IAM have not been discussed in several studies currently pertaining to RL. The use of ML in networking has also been studied in several studies [151,153]. Deep Learning (DL) methods, however, are emphasized. A summary of the known survey publications on this topic is provided in Table 4.

**Table 4.** Summary of existing survey papers.

| Study | Summary Points | Comments/Novelty |
|---|---|---|
| [80] | Industrial Internet of Things (IIoT) is evaluated from three perspectives: consensus mechanism, storage, and communication. For IIoT deployments on a large scale to be feasible, several unresolved issues must be addressed. | IoT, IDS, IPS, and IAM applications built on blockchains and RLs are examined. |
| [81] | The paper discusses a variety of cyber-physical systems, including autonomous intrusion detection algorithms, deep reinforcement learning (DRL) based game theory simulations, and multi-agent DRL simulations of cyberattack defense measures. | Reinforcement learning and cybersecurity can be viewed from three different perspectives. The major benefit of this paper is that it examines literature beyond 'Deep Reinforcement Learning'. |
| [82] | A comprehensive overview of IDS IoT ML algorithms is presented in this outstanding paper, which combines all three types of algorithms: unsupervised learning, supervised learning, and reinforcement learning. | Our research focuses on Reinforcement Learning techniques for IDS, which are used in a variety of applications, including the Internet of Things. |
| [83] | Security and privacy issues related to IoT systems are examined by the authors. A revolutionary IoT layered paradigm with specific privacy and security components is also presented. | In their study on cybersecurity concerns in IoT systems, Tawalbeh et al. [83] do not discuss machine learning-based solutions. |
| [84] | To address issues related to wireless communications and networks in the Internet of Things, this paper reviews all three classes of machine learning-based solutions at each layer of the OSI model. The feasibility of the hardware implementation is discussed. The future scope of each part is carefully examined. | A variety of application domains are reviewed in depth using RL-based approaches. |
| [85] | Analyzed machine learning workflows in the networking industry. Applications include traffic prediction, resource management, and categorization. | Their study explores the application areas of RL. Resources management and other networking-related fields have a number of applications for RL. |
| [86] | A discussion of the applications of Open Radio Access Networks (O-RAN) in supporting next-generation smart world IoT systems, providing a problem space for reliable communication, real-time analytics, fault tolerance, and interoperability. | A future research direction is outlined in the study with regard to robust and scalable solutions, interoperability and standardization, privacy, and security in O-RAN-assisted Internet of Things systems. |
| [87] | Overview of game theory in wireless communications, with a focus on security, resource allocation, power management, and spectrum usage in IoT and other wireless networks. | The purpose of this presentation is to provide an overview of game theory in wireless communications, focusing specifically on security, resource allocation, power management, and spectrum usage in IoT and other wireless networks. |
| [88] | A study of the integration of machine learning-based cognitive radio with emerging wireless | Research opportunities are explored in the study and a roadmap is provided for |

| | networks, addressing challenges such as energy efficiency, interference, throughput, latency, and security. | applying intelligent cognitive radio to next-generation wireless networks. |
|---|---|---|
| [89] | Assesses how different machine learning algorithms can be used to address issues such as security, routing, clustering, and fault detection in wireless sensor networks (WSNs). | This paper provides a comprehensive overview of ML algorithms applied to a variety of WSN issues. |
| [90] | The course discusses the use of attack graphs for the assessment of IoT vulnerabilities, including methodologies and technologies used to create and analyze attack graphs. | In this study, core modeling techniques are identified, and attack graph models are evaluated in IoT networks to provide future research directions. |
| [91] | Analyzes the detection rate, precision, and accuracy of RL-based IDS algorithms used in IDS, IPS, IoT, and IAM from 2010 to 2021. | As a result of the study, it has been identified that there are no standard evaluation criteria for real-time applications in cybersecurity, and insight is provided for new researchers in the field. |
| [92] | Analyzes the impact of machine learning techniques such as reinforcement learning, deep learning, transfer learning, and federated learning on next-generation wireless networks and the Internet of Things. | The purpose of this project is to analyze the impact of machine learning techniques on next-generation wireless networks as well as the Internet of Things, such as reinforcement learning, deep learning, transfer learning, and federated learning. |
| [93] | A survey of DRL approaches for cybersecurity is presented, covering DRL-based security methods for cyber-physical systems, autonomous intrusion detection strategies, and multi-agent DRL-based defense strategies. | DRL in cybersecurity is discussed in this study in a comprehensive manner, and future research directions are explored. |
| [94] | An examination of techniques and strategies for overcoming resource constraints in pervasive AI systems, with a focus on communication-efficient methods for distributed training and inference. | Presented in this paper are future research challenges and opportunities related to pervasive computing and artificial intelligence. |

## 5. IoT applications using blockchain technology

Blockchain and Internet of Things, Blockchain and Healthcare, Blockchain and fog computing, Blockchain and cloud computing, Blockchain and Agent, etc., were among the literature search terms used in this paper [43-47]. ACM, Springer, MDPI, SAGE and IEEE Xplore were used to gather literature from reputable databases and publishers. This paper's review of literature is presented in Figure 5.

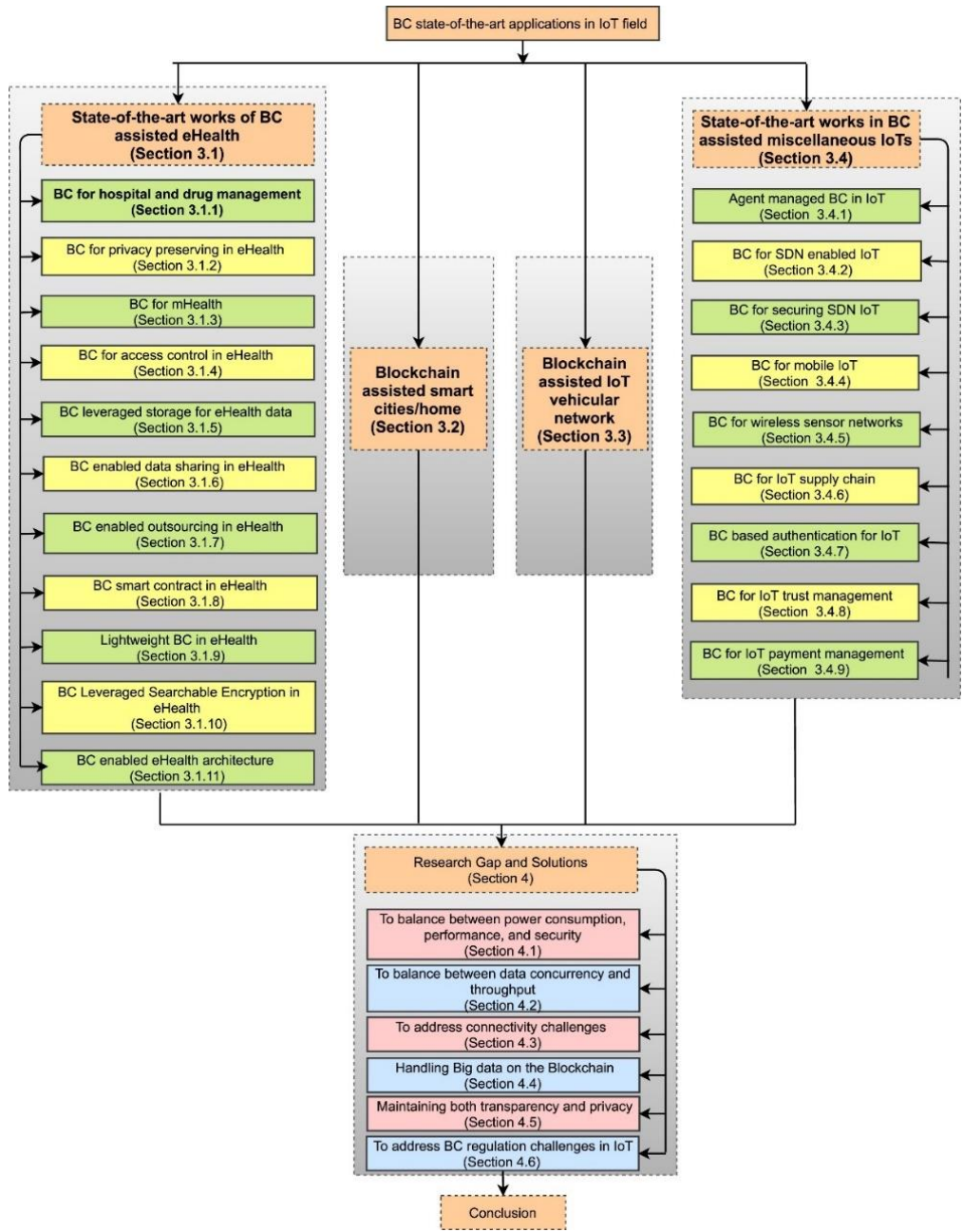

**Fig 5**. An overview of the reviewed literature is shown in the flow diagram. Blockchain refers to blockchain technology; IoT refers to the Internet of Things

## *5.1. To address regulation challenges of blockchain in IoT*

The reliability and security of blockchain make it an attractive technology for applications in finance, economics, and law. Moreover, the Congressional Cybersecurity Committee is studying blockchain technology as a possible cyberattack remedy after a recent ransomware attack on the New York Times and BBC. Despite the lack of regulation, blockchain has been used illegally in a variety of shady trade platforms, such as Silk, which is no longer operational. Due to blockchain's decentralized nature, conventional legislation cannot regulate it. Zahid et al. [48] described blockchain as a transformative mechanism from "Code is a law" to "Law is Code." Smart contracts can be used to codify laws on the blockchain. It is possible to make law a product by using blockchain smart contracts.

According to Baracaldo et al. [49], knowledge-based economies require tax, financial, and social regulation mechanisms. Blockchain and fog computing require an effective regulatory structure, they said. Effective regulation and oversight are necessary for blockchain technology to be viable and applicable. All internet applications can be adequately controlled by legislation, social norms, and economic means, according to Sun [50]. A Blockchain-based Internet of Things application can be efficiently governed by combining these four methods. The present blockchain applications are not regulated by any laws. Using IoT-oriented smart agents, social norms can be defined, and rules can be applied to blockchain applications so that they can be monitored and controlled.

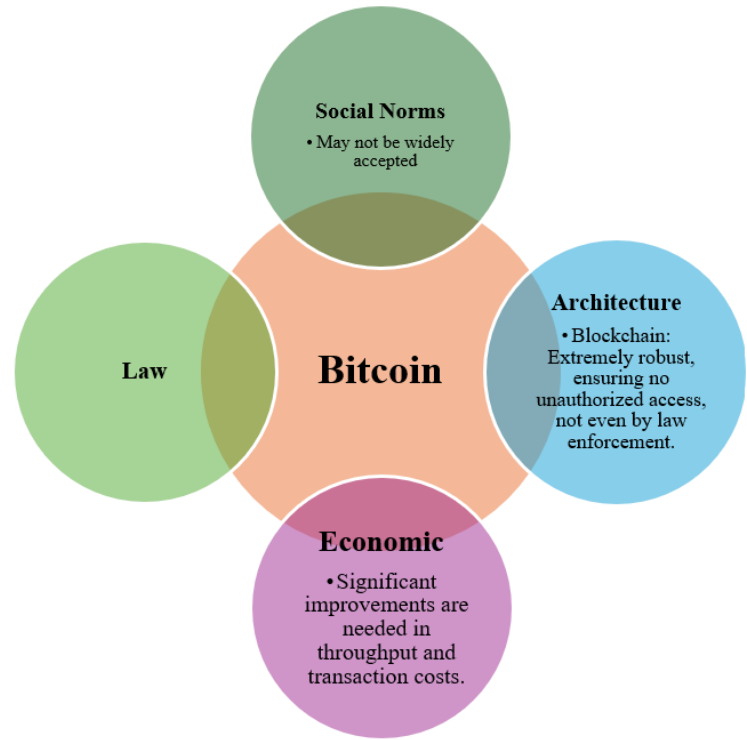

(a)

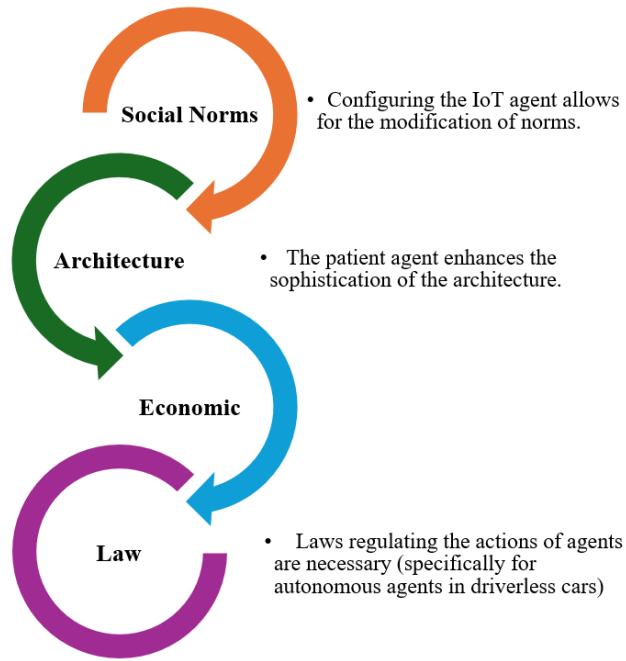

(b)

**Fig. 6.** The dot regulation of blockchain IoT (Internet of Things). Regulations in Bitcoin blockchain (a) and IoT eHealth (b).

A blockchain IoT architecture with agent-enabled technology specifies four norms of controlling technology, including social norms and regulations, that are mostly absent from existing cryptocurrencies. As shown in Figure 6 (a) and (b), Bitcoin has an extensive security architecture, but it is still in the process of implementing more regulatory mechanisms [51].

## *5.2. Access management and authentication*

It has not been extensively investigated whether RL-based authentication could be used at the application layer as well. Although PHY-layer authentication is more popular, it is also recommended that higher-level authentication methods be used. According to Yli-Huumo et al. [51], the future scope of their study will be to investigate zero-trust architectures to ensure safe authentication. It is also recommended that researchers investigate cooperative PHY-layer authentication systems in mobile WSNs and other networks of this type [52]. It has been rare to discuss the assignment of access to resources. Further investigation and assessment of the applicability and implementation of this topic are needed. As IoT and other technologies have been implemented over the last five years, RL has been investigated [52]. A few researchers are incorporating RL into existing technologies like cybersecurity. There are few articles available on authentication and real-time learning. RL and cybersecurity patents were discovered by the authors during their research for this study. China accounted for most patents. Future studies will incorporate this knowledge, according to the authors [53].

## 6. Discussion

Several benefits are associated with RL-based intelligent systems, including their cost-effectiveness, resilience, ubiquitous nature, and dynamic functionality. In closing, it is important to keep in mind the main objective of this project: identifying research topics and cutting-edge methodologies for various aspects of cybersecurity using RL. Cybersecurity has been the subject of several models and algorithms. In order to present the applications of RL algorithms in cybersecurity, we classified them into three categories. Throughout this review, we examine a wide range of IDS, IPS, IoT, and IAM reinforcement learning applications. In recent years, RL has been used in IDS classifiers and anomaly detectors. Several datasets were analyzed in this study in order to train RL-based IDS and IPS systems. We evaluated and analyzed current approaches across a variety of applications using our assessment criteria. As a result of our comprehensive analysis, we have proposed a series of open questions and suggestions for future research. By combining RL with other approaches, such as blockchain, deep neural networks, and fuzzy logic, the problem of vast, cumbersome state spaces can be solved. The result is an increase in productivity. In comparison with particle swarm optimization, RL can save 22.75% of energy for autonomous vehicle routing. In future research articles, it would be worthwhile to investigate how DRL can be used to reduce computation time and convergence time. In addition, we evaluated further polls that suggested that RL could be used to address challenges related to large-scale deployments. A zero-trust architecture for RL applications and a safe authentication method should also be investigated. Research and inquiry in reinforcement learning encompass a wide range of topics, including authentication, access management, blockchain, fuzzy logic, privacy preservation, and fuzzy heuristic evaluation. A future scope presents detailed recommendations for researchers seeking to improve the security, fail-safety, stability, intuitiveness, and human-likeness of reinforcement learning-based models. We have further categorized the existing literature on blockchain-integrated IoT systems into three main categories based on our review (papers published until 2024): Security & Privacy, Scalability and Efficiency, and Network Management. Blockchains are used to ensure secure transactions and communications in IoT networks, eliminating the need for trusted third parties.

Additionally, blockchain ensures data integrity and provenance by tracing data back through an immutable ledger and ensuring transparency and reliability. Providing robust protection against data tampering and unauthorized access, these measures address fundamental security concerns in IoT environments. Figure 7 illustrates how scalability and efficiency aim to improve resource allocation, energy management, and decentralized communication within IoT systems. As a result of blockchain technology, resources can be managed more efficiently and utilized more efficiently. A blockchain system ensures that records of energy consumption and distribution are transparent and tamper-proof. By decentralizing data processing and management, decentralized communication reduces latency and improves the scalability of IoT networks. Decentralized communication leads to more efficient and responsive systems. In order to enhance the performance and security of the Internet of Things, network management combines blockchain technology with edge and fog computing. Integrating real-time data processing closer to the data source reduces latency and improves efficiency. A fog computing approach is particularly useful in smart cities and IoT-enabled environments since it supports the decentralization of data processing and storage, thus improving the security, latency, availability, and reliability of the overall system [24-26].

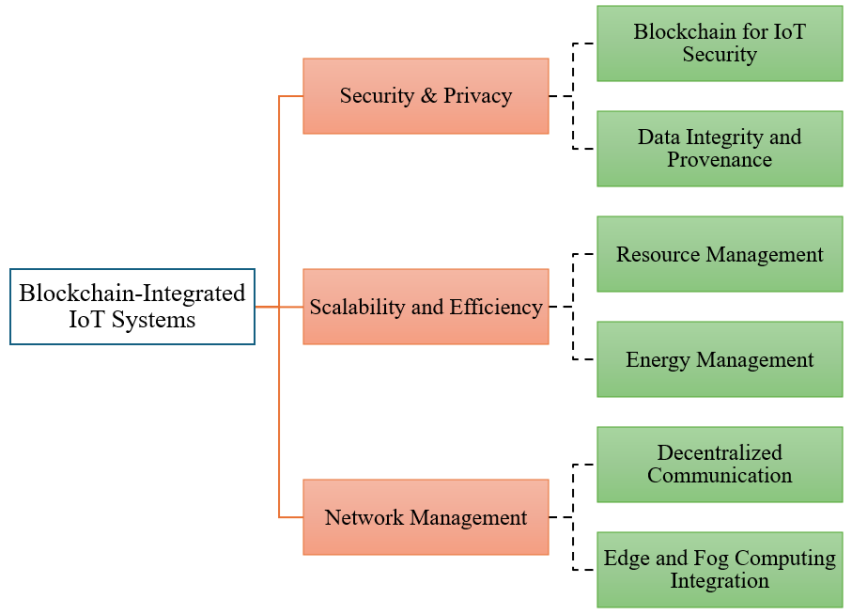

**Fig 7.** A Taxonomy for An Integrated Architecture for Blockchain and IoT-Assisted Smart Cities

A Blockchain-Integrated IoT System includes Deep Reinforcement Learning (DRL) for IoT Applications and DRL for Blockchain Networks, as shown in Figure 8. In IoT applications, DRL includes the management of autonomous IoT systems, the allocation of resources, and the management of traffic. Based on environmental changes and data inputs, DRL enables IoT systems to automatically adjust and optimize their performance. By detecting and mitigating potential threats and anomalies in real time, DRL ensures that blockchain networks are robust and secure.

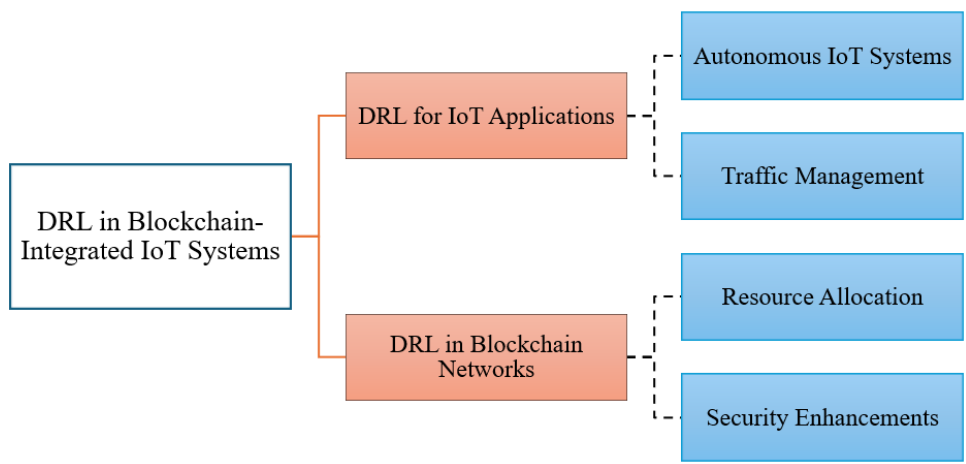

**Fig 8.** A Taxonomy for DRL Techniques in Blockchain-Enabled IoT-Assisted Smart Cities

A blockchain-integrated IoT system can be structured as an Integrated Architecture or a DRL Architecture (Figure 9). Blockchains are integrated with Internet of Things systems, cloud computing, and fog computing in order to improve scalability, security, and data processing efficiency.

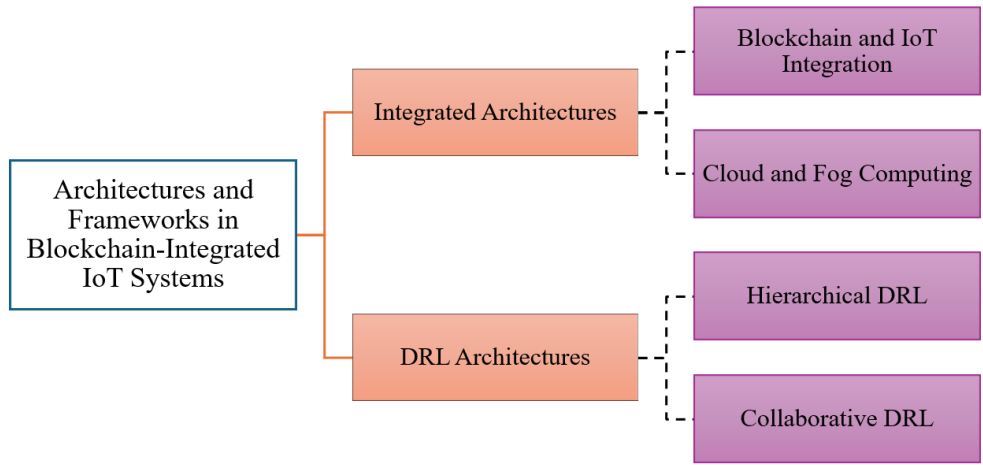

**Fig 9.** A Taxonomy for Architectures and Frameworks in Blockchain-Enabled IoT-Assisted Smart Cities

Hierarchical and collaborative DRL architectures make use of shared information and collective learning to optimize different aspects of IoT systems. The development of these advanced architectures is driving the development of more intelligent and efficient IoT applications, ensuring that they are capable of meeting the dynamic demands of smart cities.

## 6. Conclusion

An overview of DRL and blockchain collaboration is presented in this article. In our introduction to blockchain technology, we explained how this decentralized platform can help solve privacy-related challenges in machine learning. Furthermore, we discuss how blockchain can be used for DRL, as well as DRL technology and its significant applications. According to the papers we have reviewed for this study (studies from 2015 to 2024), we categorized proposed approaches and provided taxonomies to provide researchers with vital insights for future exploration and research areas. Although several challenges remain with blockchain and machine learning integration applications, the work lays the groundwork for an interdisciplinary perspective. This paper has shown the high potential of incorporating blockchain with DRL to promote IoT network services, specifically for security-sensitive and privacy-first applications. By addressing essential challenges in these important areas, namely security, privacy, data integrity, and scalability, our study provides a powerful framework that uses the immutable nature of blockchain and the adaptive capacities of DRL. Our results and conclusion from the existing literature indicate meaningful advances in decentralized decision-making, privacy-preserving data transfers, and improved mobile communication efficiency when blockchain combines with DRL in IoT-assisted smart city architectures. Also, the suggested system's capacity to cluster and classify IoT applications depicts an effective approach to addressing the sophistication of extensive IoT paradigms and ecosystems that guarantee a more secure and resilient infrastructure. In the future, we will compare the performance metrics of IoT applications with those of other applications.